# Bengali Abstractive News Summarization (BANS): A Neural Attention Approach

Prithwiraj Bhattacharjee[1][0000−0001−9300−9351], Avi Mallick[1][0000−0002−6235−0837], Md Saiful Islam[1][0000−0001−9236−380x], and Marium-E-Jannat[1][0000−0002−4815−4741]

Department of Computer Science and Engineering, Shahjalal University of Science and Technology, Sylhet-3114, Bangladesh
`prithwiraj12@student.sust.edu, avimallick999@gmail.com,`
`{saiful-cse,jannat-cse}@sust.edu`

**Abstract** Abstractive summarization is the process of generating novel sentences based on the information extracted from the original text document while retaining the context. Due to abstractive summarization's underlying complexities, most of the past research work has been done on the extractive summarization approach. Nevertheless, with the triumph of the sequence-to-sequence (seq2seq) model, abstractive summarization becomes more viable. Although a significant number of notable research has been done in the English language based on abstractive summarization, only a couple of works have been done on Bengali abstractive news summarization (BANS). In this article, we presented a seq2seq based Long Short-Term Memory (LSTM) network model with attention at encoder-decoder. Our proposed system deploys a local attention-based model that produces a long sequence of words with lucid and human-like generated sentences with noteworthy information of the original document. We also prepared a dataset of more than 19k articles and corresponding human-written summaries collected from bangla.bdnews24.com[1] which is till now the most extensive dataset for Bengali news document summarization and publicly published in Kaggle[2]. We evaluated our model qualitatively and quantitatively and compared it with other published results. It showed significant improvement in terms of human evaluation scores with state-of-the-art approaches for BANS.

## 1 Introduction

Text or document summarization is the process of transforming a long document or documents into one or more short sentences which contain the key points and main contents. Automatic summarization became vital in our daily life in order to minimize the effort and time for finding the condensed and relevant delineation of an input document that captures the necessary information of that document. Despite different ways to write the summary of a document, the summarization can be categorized into two classes based on the content selection and organization: **Extractive** and **Abstractive** approach. Extractive Summarization basically finds out the most important sentences from the text using features and grouped to produce the summary. It is like highlighting a text through a highlighter. In contrast, abstractive summarization is a technique that generates new sentences instead of selecting the essential sentences of the original document that contain the most critical information. Like a human being, writing a summary from his thinking with a pen. Machine Learning-based summarizing tools are available nowadays. But the language-specific models are hard to find.

---

[1] https://bangla.bdnews24.com/
[2] https://www.kaggle.com/prithwirajsust/bengali-news-summarization-dataset

Although a notable number of works have been done on Bengali extractive summarization, only a few abstractive summarizations are available. The majority of the available works are based on the basic Machine Learning (ML) techniques and the dataset was too small. Due to the lack of standard datasets, no significant work is available on encoder-decoder based summarization systems. So, the most challenging part for BANS is to prepare a standard and clean dataset. To build a Bengali news summarization dataset, a crawler has been made to crawl data from online resources like a daily newspaper. We have collected more than 19k data from bangla.bdnews24.com[1] online portal. The dataset represents the article and its corresponding summary.

In this paper, a sequence to sequence LSTM encoder-decoder architecture with attention has been presented for Bengali abstractive news summarization. Fig-ure 1 illustrates the proposed model. The source code and other details of the model already uploaded to Github[3]. Then the dataset of size 19096 has also been prepared which is till now the largest one and published it in Kaggle[2]. The word embedding layer has been used to represent the words in numbers and fed them into the encoder. Moreover, both the encoder and decoder parts are associated with some attention mechanisms. We got a notable improvement in terms of human assessment compared to other available Bengali abstractive summarization methods. We also evaluated ROUGE and BLEU scores. In short, our contribution to this work is threefold. They are:

– Preparation of till now the largest Bengali news summarization dataset of size 19,096 documents with its summary and published it in Kaggle[2].
– Presenting the encoder-decoder architecture with the attention mechanism for Bengali abstractive news summarization(BANS) in an efficient way.
– Evaluation of the model both qualitatively and quantitatively and the presented approach outperforms Bengali state-of-the-art approaches.

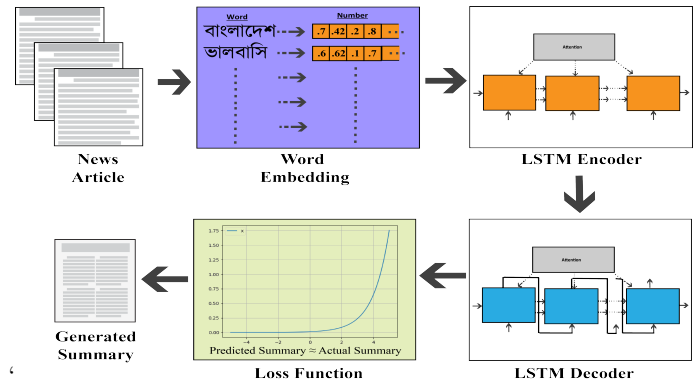

Figure 1: Illustration of our Neural Attention Model for Abstractive Summarization of Bengali News incorporates a set of LSTM encoder-decoder on top of a standard word embedding.

## 2 Related Work

There are different kinds of abstractive text summarization approaches that exist. We found that Yeasmin et al. [1] have described the different techniques regarding abstractive approaches. Then as we decided to focus on abstractive text summarization approaches on the Bengali language context, we covered Haque et al. [2] where 14 approaches of Bengali text summarization regarding both extractive and abstractive approaches are described. In 2004, Islam et al. [3] first introduced Bengali extractive summarization based on document indexing and keyword-based information retrieval. Then techniques of English extractive text summarization were applied for Bengali by Uddin et al. [4]. In 2010, Das et al. [5] used theme identification, page rank algorithms, etc. for extractive summarization. Sentence ranking and stemming process-based Bengali extractive summarization were first proposed by a researcher named Kamal Sarkar [6] and later in a better way by Efat et al. [7].
Haque et al. [8,9] respectively proposed a key-phrase based extractive approach and a pronoun replacement based sentence ranking approach. In 2017, the heuristic approach proposed by Abujar et al. [10], K-means clustering method of Akther et al. [11] and LSA (Latent Semantic Analysis) method stated in Chowdhury et al. [12] became popular techniques for Bengali extractive summarization. The graph-based sentence scoring feature for Bengali summarization was first used by Ghosh et al. [13]. Moreover, Sarkar et al. [14] and Ullah et al. [15] proposed term frequency and cosine similarity based extractive approach respectively.
Recently, Munzir et al. [16] instigated a deep neural network-based Bengali extractive summarization. Again Abujar et al. [17] introduced Word2Vec based word embedding for Bengali text summarization. Then Talukder et al. [18]

---
[3] https://github.com/Prithwiraj12/Bengali-Deep-News-Summarization

proposed an abstractive approach for Bengali where bi-directional RNNs with LSTM are used at the encoder and attention at the decoder. We also used LSTM-RNN based attention model like [18] but we applied attention to both the encoder and the decoder layer and did some comparative study with the corresponding result part and dataset part with the existing one. Another LSTM-RNN based text generation process is introduced by Abujar et al. [19] for Bengali abstractive text summarization.

We used the concept stated in Lopyrev et al. [20] for our system. The seq2seq model and the LSTM encoder-decoder architecture we used, was introduced by Sutskever et al. [21] and Bahdanau et al. [22] respectively. Again, the decoder and encoder part's attention technique is the concept stated in Luong et al. [23] and Rush et al. [24] respectively. Furthermore, the LSTM concept-based language parsing method has been adopted from Vinyals et al. [25].

## 3 Dataset

A standard dataset is a vital part of text summarization. We gathered a conceptual idea of preparing a standard dataset from Hermann et al. [26] and also observed some of the existing public English datasets like CNN-Daily Mail[4] dataset. We need a vast amount of data for training but no significant standard public dataset is available for Bengali summarization. So, we collected

Table 1: Statistics of the dataset

| Total No of Articles | 19,096 |
|---|---|
| Total No of Summaries | 19,096 |
| Maximum No of Words in an Article | 76 |
| Maximum No of Words in a Summary | 12 |
| Minimum No of Words in an Article | 5 |
| Minimum No of Words in a Summary | 3 |

news and its summary from the online news portal bangla.bdnews24.com[1] as it had both the article and its summary. We made a crawler and crawled 19352 news articles and their summaries from different categories like sports, politics, economics, etc. Online news contains lots of garbage like advertisements, non-Bengali words, different websites' links, etc. So, we started preprocessing by making a data cleaning program that eliminates all kinds of garbage from the dataset. We uploaded data crawling, cleaning, and analysis source code[5] and their working details to Github and publicly published our dataset in Kaggle[2]. A tabular representation of our processed data is shown in Table 1. The significance and comparison of our dataset with only publicly available Bangla Natural

---

[4] https://cs.nyu.edu/ kcho/DMQA/
[5] https://github.com/Prithwiraj12/Data-Manipulation

Language Processing Community (BNLPC[6]) summarization dataset has been shown in Table 2.

Table 2: Comparison of our standard dataset with BNLPC dataset

| Source | Total Articles | No of summary (per article) | Total Summaries |
| --- | --- | --- | --- |
| BNLPC[6] Dataset | 200 | 3 | 600 |
| Our Dataset[2] | 19096 | 1 | 19096 |

# 4  Model Architecture

By observing the significant performance of LSTM encoder-decoder with the attention mechanism described in Lopyrev et al. [20], we've used a similar neural attention model architecture. It has an LSTM Encoder part and an LSTM Decoder part. Both of the parts are associated with some attention mechanisms. Tensorflow's embedding layer embedding_attention_seq2seq has been used to represent the words in numbers to feed into encoders. After generating the decoder's output, a comparison between the actual and predicted summary has been done using the softmax loss function, and for minimizing the loss, the network started back-propagating. Lastly, a summary has been generated with minimal loss. The whole process works as a seq2seq approach and can be visualized by figure 1. Let's describe the major two components of our model.

Firstly, an input sequence is encoded to numbers via word embedding layer and fed into the LSTM encoder in reverse order. Sutskever et al. [21] proposed that because of calculating short term dependencies, the first few words of both the input sequence and output sequence must be closer to each other and it can be achieved by feeding input in reverse order and thus the result can be significant. That means Bengali sentence like "আজকের সংবাদ" is fed into each encoder cell reversely as individual word "সংবাদ" and "আজকের" respectively. Attention is also used to the encoder part as mentioned by Rush et al. [24].

Secondly, we used a greedy LSTM decoder which is different from a beam search decoder. Firstly, encoder output is fed into the first decoder cell. Then the output of the current decoder cell is fed into the next decoder cell along with the attention as well as the information from the previous decoder cell and continued the process till the last decoder cell. That means if the first generated word in the decoder cell is "সংবাদের" then this word will help to predict the next word suppose "সারাংশ" for the next decoder cell combining with attention and continued the process till the end. The decoder attention mechanism is implemented as stated in [21].

Before training, we made a vocabulary of the most frequent 40k words both from articles and summaries. The out of vocabulary words are denoted by _UNK token. _PAD token is used for padding the article and its summary to the bucket sizes. A bucket is nothing but an array where we define how many words an article and its summary can hold while training. We used five encoder-decoder LSTM models for training. Now, the trained model also padded the words of the given input sentences to the bucket sizes. So the model can well summarize the articles containing the number of words in all sentences equal to the largest bucket size and in our case it was (50, 20) for article and summary respectively.

---

[6] http://www.bnlpc.org/research.php

## 5 Result and Discussion

We assessed our model based on two types of evaluation matrices for analyzing the result: They are **Quantitative Evaluation** and **Qualitative Evaluation**. Both of the evaluation methods are mandatory for checking how much the summary system is suitable for generating a summary. 70% of our data was used for training, 20% for validating, and 10% was used for testing. The system was trained three times with different parameter specifications. After the evaluation, we found that the system has the best output when the vocabulary size was set to 40k, hidden unit to 512, learning rate to 0.5, and steps per checkpoint to 350. Table 3 shows some generated examples of our best model. We showed two good quality as well as two poor quality predictions in table 3 from our system. Here, the first two predictions are well summarised by our model and sometimes the new word has also been generated like "পুকুরে" in the second example. On the other hand, from the last two predictions on the table 3 we found that repetition of words like "দগ্ধ" in the third example and "লাশ" in the fourth example occurred twice. Further from the third example, we can see inaccurate reproduction of factual details. That means word "কুষ্টিয়া" has been produced by the model rather than predicting the word "ঠাকুরগাঁও" in the fourth example. Moreover, due to bucketing issues, some summaries are forcefully stopped before hitting the end token of the sentence which can be shown in third predictions on table 3.

### 5.1 Quantitative Evaluation

Quantitative evaluation is a system-oriented evaluation. In this evaluation process, both the actual and predicted summaries are given as input to a program and the program generates a score comparing how much the predicted summary deviates from the actual summary. We found that Recall-Oriented Understudy for Gisting Evaluation (ROUGE) [27] and Bilingual Evaluation Understudy (BLEU) [28] are two standard quantitative evaluation matrices. As far as our knowledge, quantitative evaluation of the existing Bengali abstractive text summarization techniques [18, 19] is not mentioned or publicly available. So we could not compare our evaluation with them. But as per standard scoring mentioned in the papers [27, 28], our achieved score was also significant. There are different variants of ROUGE calculation exist. ROUGE-1, ROUGE-2, ROUGE-L, ROUGE-N, etc are some of them. Here, we computed the most adapted ROUGE-1, ROUGE-L, and measured the BLEU score as well. Firstly, We took 100 generated summaries and corresponding actual summaries and calculated the average BLEU score. Again for ROUGE calculation, we first calculated the Precision and Recall. Then using these two measurements calculated the average F1 score for that 100 examples. The bar diagram of figure 2 denotes ROUGE and BLEU scores of the best model.

Table 3: Illustrates some predictions of our BANS system showing the input news article, actual summary and BANS predicted summary

| New Article | Actual Summary | Predicted Summary |
|---|---|---|
| **Bengali:** বিএনপি জোটের হরতাল-অবরোধের মধ্যে চট্টগ্রাম নগরীর স্টিল মিল এলাকায় একটি বাসে আগুন দিয়েছে দুর্বৃত্তরা।<br><br>**English:** The miscreants set fire to a bus in Steel Mill area of Chittagong city during the strike and blockade of the BNP alliance. | **Bengali:** চট্টগ্রামে বাসে আগুন।<br><br>**English:** Fire on bus in Chittagong. | **Bengali:** চট্টগ্রামে বাসে বাসে আগুন<br><br>**English:** Fire on the buses of Chittagong |
| **Bengali:** জামালপুরের মেলান্দহে পানিতে ডুবে দুই শিশুর মৃত্যু হয়েছে।<br><br>**English:** Two children drowned at Melandho in Jamalpur. | **Bengali:** জামালপুরে পানিতে ডুবে দুই শিশুর মৃত্যু<br><br>**English:** Two children drowned in Jamalpur | **Bengali:** পুকুরে পানিতে ডুবে শিশুর মৃত্যু<br><br>**English:** The child drowned in the pond |
| **Bengali:** নরসিংদীর পলাশে অবরোধের পেট্রোলবোমায় দগ্ধ হওয়ার আট দিন পর হাসপাতালে চিকিৎসাধীন অবস্থায় মারা গেলেন ট্রাকচালক জাহিদ আহমেদ।<br><br>**English:** Truck driver Zahid Ahmed died in a hospital after eight days being burnt by a petrol bomb in a blockade at Polash in Narsingdi. | **Bengali:** নরসিংদীতে বোমায় দগ্ধ ট্রাকচালকের মৃত্যু<br><br>**English:** Truck driver killed in bomb blast in Narsingdi | **Bengali:** অবরোধের দগ্ধ দগ্ধ আরও মৃত্যু<br><br>**English:** More death from burns burns in the siege |
| **Bengali:** ঠাকুরগাঁও সদরে নদীতে পড়ে নিখোঁজের একদিন এক যুবকের লাশ উদ্ধার করা হয়েছে।<br><br>**English:** The body of a young man who went missing after falling into the river at Thakurgaon Sadar has been recovered. | **Bengali:** ঠাকুরগাঁওয়ে নিখোঁজ যুবকের লাশ উদ্ধার।<br><br>**English:** The dead body of a missing young boy was recovered in Thakurgaon. | **Bengali:** কুষ্টিয়ায় নিখোঁজ যুবকের লাশ লাশ উদ্ধার<br><br>**English:** The dead body dead body of a missing young boy was recovered in Kushtia |

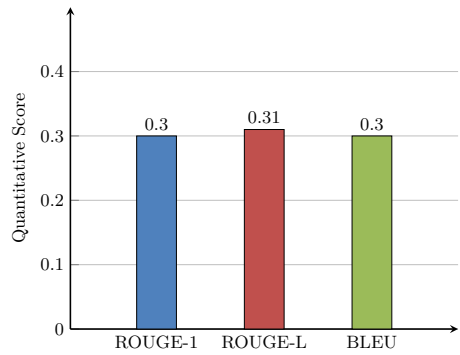

Figure 2: Illustrates the Quantitative analysis of our proposed model based on ROUGE-1, ROUGE-L and BLEU scores

### 5.2 Qualitative Evaluation

Qualitative evaluation is the user-oriented evaluation process. Here some users of different ages take part in rating the generated summary on a scale of 5 compared with the actual one. For the qualitative evaluation, we took some examples from our system and some from the existing one [18]. As far as our knowledge, qualitative evaluation of the existing method [18] is not publicly available. So for comparison, we also had to calculate the rating for [18]. We provided the examples of both the systems to the users via a google form[7] survey. A total of 20 users participated in a rating on a scale of 5. Among the users 45% were female and 55% were male. Moreover, all the users were from the educational background with an average age of 24. Again 45% were from linguistic faculty, 35% were from engineering faculty and 25% were from other faculties. We calculated the average rating regarding each of the models and found that our system outperforms the existing system based on human assessment. The qualitative rating of the systems is shown in table 4.

Table 4: Qualitative evaluation of existing system and the proposed system

| System | Average Rating(Out of 5) |
|:---:|:---:|
| Proposed System | 2.80 |
| Existing System [18] | 2.75 |

## 6  Conclusion

To recapitulate, the development of the standard summarization dataset of 19,096 Bengali news has been one of our pioneering accomplishments, especially since it is the largest publicly published dataset in this field. Here a neural attention-based encoder-decoder model for abstractive summarization of Bengali news has been presented, which generates human-like sentences with core information of the original documents. Along with that, a large-scale experiment was conducted to investigate the effectiveness of the proposed BANS. From the qualitative evaluation, we have found that the proposed system generates more humanoid output than all other existing BANS. Indeed, the LSTM-based encoder-decoder has been exceptionally successful, nonetheless, the model's performance can deteriorate quickly for long input sequences. Repetition of summaries and inaccurate reproduction of factual details are two significant problems. To fix these issues, we plan to drive our efforts on modeling hierarchical encoder based on structural attention or pointer-generator architecture and developing methods for multi-document summarization.

---

[7] https://forms.gle/r9Mu5NEpVkMcSXbD9

**Acknowledgements**
We would like to thank Shahjalal University of Science and Technology (SUST) research center and SUST NLP research group for their support.

## References

1. Yeasmin, S., Tumpa, P.B., Nitu, A.M., Uddin, M.P., Ali, E., Afjal, M.I.: Study of abstractive text summarization techniques. American Journal of Engineering Research **6**(8), 253–260 (2017)
2. Haque, M.M., Pervin, S., Hossain, A., Begum, Z.: Approaches and trends of automatic bangla text summarization: Challenges and opportunities. International Journal of Technology Diffusion (IJTD) **11**(4), 1–17 (2020)
3. Islam, M.T., Al Masum, S.M.: Bhasa: A corpus-based information retrieval and summariser for bengali text. In: Proceedings of the 7th International Conference on Computer and Information Technology (2004)
4. Uddin, M.N., Khan, S.A.: A study on text summarization techniques and implement few of them for bangla language. In: 2007 10th international conference on computer and information technology. pp. 1–4. IEEE (2007)
5. Das, A., Bandyopadhyay, S.: Topic-based bengali opinion summarization. In: Coling 2010: Posters. pp. 232–240 (2010)
6. Sarkar, K.: Bengali text summarization by sentence extraction. arXiv preprint arXiv:1201.2240 (2012)
7. Efat, M.I.A., Ibrahim, M., Kayesh, H.: Automated bangla text summarization by sentence scoring and ranking. In: 2013 International Conference on Informatics, Electronics and Vision (ICIEV). pp. 1–5. IEEE (2013)
8. Haque, M.M., Pervin, S., Begum, Z.: Enhancement of keyphrase-based approach of automatic bangla text summarization. In: 2016 IEEE Region 10 Conference (TENCON). pp. 42–46. IEEE (2016)
9. Haque, M., Pervin, S., Begum, Z., et al.: An innovative approach of bangla text summarization by introducing pronoun replacement and improved sentence ranking. Journal of Information Processing Systems **13**(4) (2017)
10. Abujar, S., Hasan, M., Shahin, M., Hossain, S.A.: A heuristic approach of text summarization for bengali documentation. In: 2017 8th International Conference on Computing, Communication and Networking Technologies (ICCCNT). pp. 1–8. IEEE (2017)
11. Akter, S., Asa, A.S., Uddin, M.P., Hossain, M.D., Roy, S.K., Afjal, M.I.: An extractive text summarization technique for bengali document (s) using k-means clustering algorithm. In: 2017 IEEE International Conference on Imaging, Vision & Pattern Recognition (icIVPR). pp. 1–6. IEEE (2017)

10      P. Bhattacharjee et al.12. Chowdhury, S.R., Sarkar, K., Dam, S.: An approach to generic bengali text summarization using latent semantic analysis. In: 2017 International Conference on Information Technology (ICIT). pp. 11–16. IEEE (2017)
13. Ghosh, P.P., Shahariar, R., Khan, M.A.H.: A rule based extractive text summarization technique for bangla news documents. International Journal of Modern Education and Computer Science **10**(12), 44 (2018)
14. Sarkar, A., Hossen, M.S.: Automatic bangla text summarization using term frequency and semantic similarity approach. In: 2018 21st International Conference of Computer and Information Technology (ICCIT). pp. 1–6. IEEE (2018)
15. Ullah, S., Hossain, S., Hasan, K.A.: Opinion summarization of bangla texts using cosine simillarity based graph ranking and relevance based approach. In: 2019 International Conference on Bangla Speech and Language Processing (ICBSLP). pp. 1–6. IEEE (2019)
16. Al Munzir, A., Rahman, M.L., Abujar, S., Hossain, S.A., et al.: Text analysis for bengali text summarization using deep learning. In: 2019 10th International Conference on Computing, Communication and Networking Technologies (ICCCNT). pp. 1–6. IEEE (2019)
17. Abujar, S., Masum, A.K.M., Mohibullah, M., Hossain, S.A., et al.: An approach for bengali text summarization using word2vector. In: 2019 10th International Conference on Computing, Communication and Networking Technologies (ICCCNT). pp. 1–5. IEEE (2019)
18. Talukder, M.A.I., Abujar, S., Masum, A.K.M., Faisal, F., Hossain, S.A.: Bengali abstractive text summarization using sequence to sequence rnns. In: 2019 10th International Conference on Computing, Communication and Networking Technologies (ICCCNT). pp. 1–5. IEEE (2019)
19. Abujar, S., Masum, A.K.M., Islam, M.S., Faisal, F., Hossain, S.A.: A bengali text generation approach in context of abstractive text summarization using rnn. In: Innovations in Computer Science and Engineering, pp. 509–518. Springer (2020)
20. Lopyrev, K.: Generating news headlines with recurrent neural networks. arXiv preprint arXiv:1512.01712 (2015)
21. Sutskever, I., Vinyals, O., Le, Q.V.: Sequence to sequence learning with neural networks. In: Advances in neural information processing systems. pp. 3104–3112 (2014)
22. Bahdanau, D., Cho, K., Bengio, Y.: Neural machine translation by jointly learning to align and translate. arXiv preprint arXiv:1409.0473 (2014)
23. Luong, M.T., Pham, H., Manning, C.D.: Effective approaches to attention-based neural machine translation. arXiv preprint arXiv:1508.04025 (2015)
24. Rush, A.M., Chopra, S., Weston, J.: A neural attention model for abstractive sentence summarization. arXiv preprint arXiv:1509.00685 (2015)
25. Vinyals, O., Kaiser, Ł., Koo, T., Petrov, S., Sutskever, I., Hinton, G.: Grammar as a foreign language. In: Advances in neural information processing systems. pp. 2773–2781 (2015)
26. Hermann, K.M., Kocisky, T., Grefenstette, E., Espeholt, L., Kay, W., Suleyman, M., Blunsom, P.: Teaching machines to read and comprehend. In: Advances in neural information processing systems. pp. 1693–1701 (2015)
27. Lin, C.Y.: Rouge: A package for automatic evaluation of summaries. In: Text summarization branches out. pp. 74–81 (2004)
28. Pastra, K., Saggion, H.: Colouring summaries bleu. In: Proceedings of the EACL 2003 Workshop on Evaluation Initiatives in Natural Language Processing: are evaluation methods, metrics and resources reusable? pp. 35–42 (2003)